\documentclass[letterpaper,10pt,conference]{ieeeconf}
\IEEEoverridecommandlockouts       
\usepackage{amssymb}
\usepackage{amsmath}
\usepackage[english]{babel}
\usepackage{float}
\usepackage{graphicx}
\usepackage{hyperref}
\usepackage{mathtools}

\usepackage{filecontents}
\usepackage[noadjust]{cite}
\usepackage[font=footnotesize,labelfont=footnotesize]{subcaption}
\usepackage[font=footnotesize,labelfont=footnotesize]{caption}
\usepackage{comment}
\usepackage{authblk}
\usepackage{multirow}
\usepackage{xcolor}
\usepackage{algorithm}
\usepackage{algorithmic}
\usepackage{xspace}
\usepackage{graphicx}

\DeclareMathOperator*{\argmin}{arg\,min}

\definecolor{darkblue}{rgb}{0.15,0.15,0.55}
\definecolor{lightgrey}{rgb}{0.75,0.75,0.75}

\begin{document}
\title{\LARGE \bf Escaping Local Minima: Hybrid Artificial Potential Field with Wall-Follower for Decentralized Multi-Robot Navigation\vspace{-5pt}}
\author{Joonkyung Kim$^1$, Sangjin Park$^1$,  Wonjong Lee$^1$, Woojun Kim$^2$, Nakju Doh$^3$, and Changjoo Nam$^{1,*}$
\thanks{This work was supported in part by Samsung Research Funding \& Incubation Center of Samsung Electronics under Project Number SRFC-TD2103-01 and the National Research Foundation of Korea (NRF) grant funded by the Korea government (MSIT) (No. RS-2024-00411007). $^1$Dept. of Electronic Engineering, Sogang University, South Korea. $^2$Robotics Institute, Carnegie Mellon University, USA. $^3$TeeLabs, South Korea. $^*$Corresponding author: {\tt\small cjnam@sogang.ac.kr}}
}

\maketitle

\begin{abstract}
We tackle the challenges of decentralized multi-robot navigation in environments with nonconvex obstacles, where complete environmental knowledge is unavailable. While reactive methods like Artificial Potential Field (APF) offer simplicity and efficiency, they suffer from local minima, causing robots to become trapped due to their lack of global environmental awareness. Other existing solutions either rely on inter-robot communication, are limited to single-robot scenarios, or struggle to overcome nonconvex obstacles effectively.

Our proposed methods enable collision-free navigation using only local sensor and state information without a map. By incorporating a wall-following (WF) behavior into the APF approach, our method allows robots to escape local minima, even in the presence of nonconvex and dynamic obstacles including other robots. We introduce two algorithms for switching between APF and WF: a rule-based system and an encoder network trained on expert demonstrations. Experimental results show that our approach achieves substantially higher success rates compared to state-of-the-art methods, highlighting its ability to overcome the limitations of local minima in complex environments.

\end{abstract}

\section{Introduction}
\label{sec:intro}

Multi-robot navigation presents challenges owing to the need for coordination and collision avoidance with both static and dynamic obstacles, including other robots. In environments with limited wireless communication, such as complex indoor settings~\cite{mcguire2019science} or expansive outdoor spaces~\cite{bayer2023subterranean}, centralized methods face issues like single-point failures and incomplete state information, making them unreliable in large, dynamic environments.

Decentralized approaches, such as Optimal Reciprocal Collision Avoidance (ORCA)~\cite{van2008orca} and Control Barrier Functions (CBF)~\cite{ames2019cbf}, offer an alternative by using local sensor data. However, these methods can experience deadlocks, where robots become stuck due to conflicting objectives between goal-seeking and collision avoidance~\cite{grover2023deadlock}. Additionally, ORCA and CBF often require local maps and the states of nearby robots, which can be computationally expensive or difficult to obtain~\cite{qin2023srl_orca,gao2023onlineCBF}.

\begin{figure}[t]
    \captionsetup{skip=0pt}
    \centering
    \begin{subfigure}{0.42\textwidth}
        \captionsetup{skip=0pt}
        \centering
	   \includegraphics[width=\textwidth]{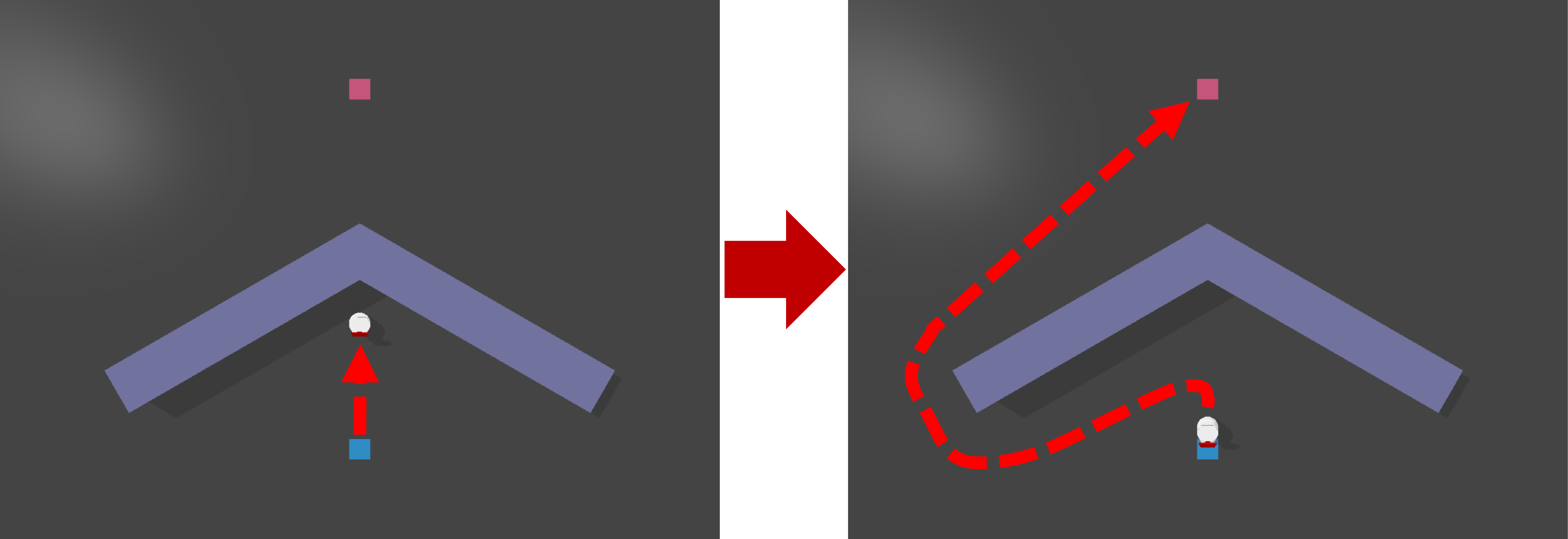}
	   \caption{Local minimum trap caused by a nonconvex obstacle}
        \label{fig:1_a}
    \end{subfigure}
    \hfill
    \begin{subfigure}{0.42\textwidth}
        \centering
        \captionsetup{skip=0pt}
    	\includegraphics[width=\textwidth]{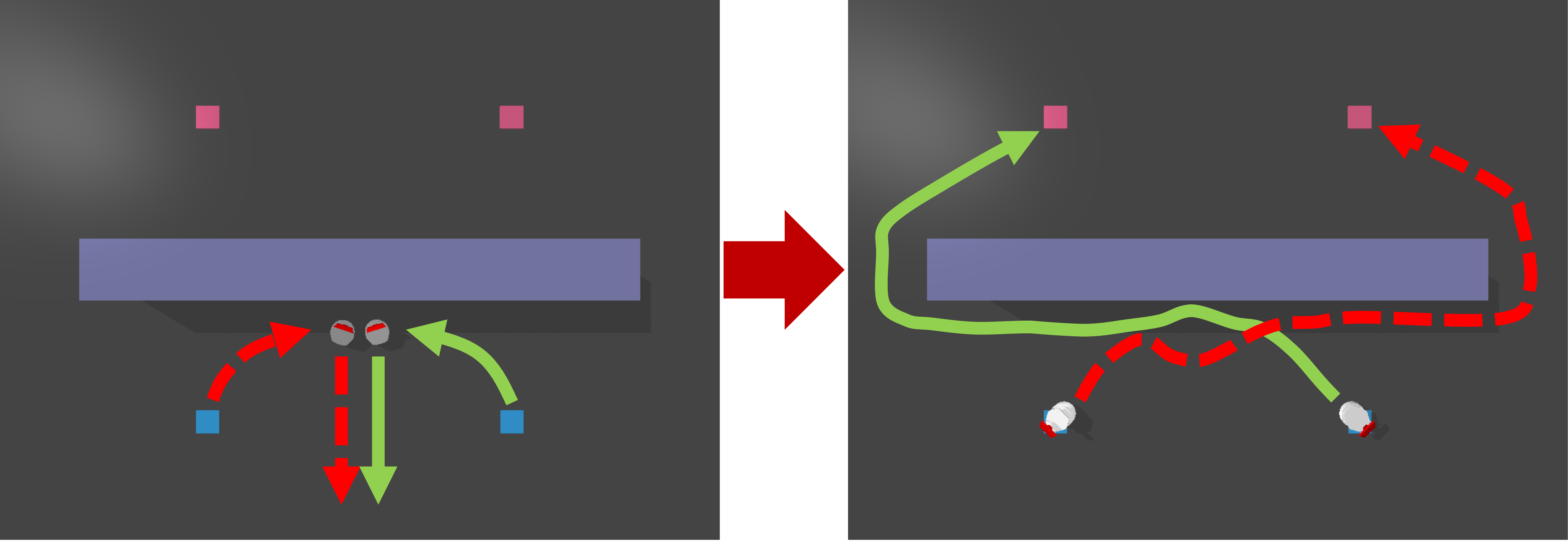}
    	\caption{Robots misinterpreting each other as static obstacles}
        \label{fig:1_b}
    \end{subfigure}    
    \caption{Challenges in decentralized navigation with limited environmental information. The problematic paths of robots due to incomplete environmental data (left) which can be overcame using our proposed method (right).}
    \label{fig:dummy1}\vspace{-22pt}
\end{figure}

Mapless navigation is ideal when building a common map is impractical. A well-known method for mapless navigation is the Artificial Potential Field (APF)~\cite{apf}, commonly used in multi-robot systems~\cite{yu2021smmrAPFexplore,2023rpf}. However, APF suffers from the local minima problem, where a robot can become trapped between the attractive force of its goal and the repulsive force from obstacles~\cite{yun1997APFlocalMin}, especially in nonconvex environments as shown in Fig.~\ref{fig:1_a}.

A common approach to overcome local minima is to combine APF with Wall-Following (WF). WF is a reactive method that allows robots to detect and follow obstacle boundaries without complete environmental information~\cite{lumelsky1986bug1_bug2}. When combined with APF, WF helps robots escape local minima, enabling them to handle complex, nonconvex environments without any map~\cite{borenstein1989realtime, yun1997wall_icar, zhu2009improvedWF, 2008uNAV, 2023rpf, 2024dacoop}. However, many APF-WF methods are designed for single-robot scenarios and do not account for dynamic obstacles~\cite{borenstein1989realtime, yun1997wall_icar, zhu2009improvedWF, 2008uNAV}. In multi-robot systems, WF often requires perfect communication~\cite{sarid2007mrbug, johansson2023swarmBug, mcguire2019science} or imposes restrictive conditions such as one-directional WF~\cite{tan2023OA_Bug_Swarm} or convex obstacles~\cite{2023rpf, 2024dacoop}.

This work tackles decentralized, multi-robot navigation in environments with limited information. Each robot navigates toward its goal without a pre-built map, geometric information of obstacles, or communication with other robots. We propose a decentralized APF-WF system using two switching schemes: (i) a Rule-based Switch (RS) that triggers wall-following based on local sensor data, and (ii) a Learned Switch (LS) that enhances RS using a Vision Transformer (ViT) encoder trained on human expert decisions. RS helps robots escape local minima, while LS reduces occurrences where robots move in parallel as they recognize each other as a wall to follow, as shown in Fig.~\ref{fig:1_b}. 

The key contributions of this work include the development of RS that enables decentralized, mapless navigation in dynamic and nonconvex environments, the introduction of a deep encoder network LS to leverage human decision-making, an extensive comparison with other methods, and an experiment with physical robots.

\section{Related Work}
\label{sec:relWork}
\vspace{-3pt}

APF is a popular reactive navigation strategy that guides robots toward their targets using local sensor data without needing a complete map. Its simplicity and efficiency make it widely used in tasks like exploration, path planning, and cooperative pursuit. However, it is prone to local minima~\cite{yun1997APFlocalMin}, especially in nonconvex environments, where robots can become trapped between attractive and repulsive forces.

WF is another reactive method, often used for navigating unknown spaces by detecting and following boundaries. Early algorithms like Bug1 and Bug2~\cite{lumelsky1986bug1_bug2} ensure completeness for single robots in static environments, while TangentBug~\cite{kamon1998tangentbug} needs precise geometric information of obstacles. MRBug~\cite{sarid2007mrbug} extends WF to multi-robot systems but relies on inter-robot communication, making it unsuitable for decentralized approaches.
Recent algorithms, such as Swarm Gradient Bug~\cite{mcguire2019science} and Swarm Bug~\cite{johansson2023swarmBug}, have applied WF in multi-robot navigation without detailed maps but still depend on communication. While OA Bug~\cite{tan2023OA_Bug_Swarm} combines WF with olfactory and auditory sensors not to rely on a map and communication, its reliance on a fixed WF direction for conflict avoidance reduces efficiency.

Combining APF and WF is common to overcome local minima in single-robot navigation. An early work~\cite{borenstein1989realtime} introduces APF-WF switching for obstacle avoidance but could fall into infinite loops of switching. Later, \cite{yun1997wall_icar} and \cite{zhu2009improvedWF} improve handling of complex nonconvex obstacles using local observations and simple memory. However, they are limited to single-robot scenarios. $\mu$NAV~\cite{2008uNAV} dynamically switches between APF and WF by adjusting switching weights. While it provides theoretical completeness in nonconvex environments, its applicability remains confined to single-robot static settings.
Recent developments like RPF~\cite{2023rpf} and DACOOP-A~\cite{2024dacoop} extend APF-WF strategies to multi-robot systems. While they incorporate WF to escape local minima and manage inter-robot conflicts, they still struggle in nonconvex environments and complex real-world scenarios.

\vspace{-3pt}
\section{Problem Description}
\label{sec:prob_desc}
\vspace{-3pt}

We consider the problem of decentralized navigation of $N$ robots, each tasked with reaching a goal within a finite time $T$. The robots operate in an unknown environment, relying on local sensing to avoid obstacles and each other without a prior map and communication. Our objective is to enable all robots to complete their tasks without any collision.

Each robot $R_i$ ($i \in \{1, 2, \ldots, N\}$) with radius $r$ operates in a 2D workspace $\mathcal{W} \subset \mathbb{R}^2$ that contains unknown obstacles $ \mathcal{W}_{\text{obs}} $ and free space $ \mathcal{W}_{\text{free}} = \mathcal{W} \setminus \mathcal{W}_{\text{obs}}$. The state of each robot at time $t$ can be described by its position and orientation (or heading angle), denoted by $\mathbf{s}_i(t)=[{x}_i(t), {y}_i(t), {\psi}_i(t)]$ where the position $[{x}_i(t), {y}_i(t)]$ forms $\mathbf{p}_i(t)$ and the heading angle ${\psi}_i(t)$ is in range $(-\pi, \pi]$. Each robot must navigate from its initial position $\mathbf{p}_i(0)$ to its goal $\mathbf{g}_i= [{x}_{\text{goal}, i}, {y}_{\text{goal}, i}]$ while avoiding collisions where the time at $\mathbf{g}_i$ is denoted by $t_i$. 
For practical implementation of robot navigation, we consider the task of $R_i$ to be successful if $R_i$ reaches within a small tolerance $\epsilon > 0$ of $\mathbf{g}_i$ (i.e., $\parallel\mathbf{p}_i(t_i) - \mathbf{g}_i\parallel \leq \epsilon$) with $t_i  \in [0, T]$.

We assume that $R_i$ is given its state information $\mathbf{s}_i(t)$ and the relative target position $\mathbf{g}_{\text{rel},i}(t) \in \mathbb{R}^2$ from $\mathbf{p}_i(t)$. Also, we assume that robots are equipped with a 360-degree range sensor (e.g., 2D LiDAR) with the maximum range $d_\text{max}$. Each robot estimates its pose within its reference frame (denoted as \textit{Initial Frame ($\text{IF}$)}) based on its initial pose. The goal is also represented in IF. Robots have a constant-sized memory to store a sequence of historical records.

Sensor data processing and navigation are performed in the internal local frame of robot which is called \textit{Robot Frame (RF)}. In RF, the robot heading is aligned with the positive $x$-axis, while the positive $y$-axis is aligned with the direction to the right side of the robot. To denote the reference frame in calculations, we use IF as a superscript, while RF is left implied (i.e., without a superscript). For instance, the current state in the initial frame is $\mathbf{x}^{\text{IF}}_{i}(t) = [x^{\text{IF}}_i(t), y^{\text{IF}}_i(t), \psi^{\text{IF}}_i(t)]$. If necessary, we indicate the individual coordinate system (IF and RF) of $R_i$ using the subscript $i$.

\vspace{-3pt}
\section{Preliminary: Artificial Potential Fields}
\label{sec:prelim}
\vspace{-3pt}

APF aims to guide a robot towards a goal while avoiding obstacles, based on data from an omnidirectional range sensor. Each robot calculates the attractive force $\mathbf{F}_{\text{att}}$ to move toward the goal and the repulsive force $\mathbf{F}_{\text{rep}}$ to avoid obstacles. The goal vector in RF (i.e., $\mathbf{g}_{\text{rel}}(t)$) is computed based on its goal $\mathbf{g}^{\text{IF}}$ and the current relative state $\mathbf{s}^{\text{IF}}(t)$ measured from the odometry. The force $\mathbf{F}_{\text{att}}(\mathbf{g}_{\text{rel}}(t)) \in \mathbb{R}^2$ is \vspace{-4pt}
\begin{equation}
\small{
    \mathbf{F}_{\text{att}}(\mathbf{g}_{\text{rel}}(t)) =  \triangledown U_{\text{att}}(\mathbf{g}_{\text{rel}}(t)) = \mathbf{g}_{\text{rel}}(t), \label{eq:att_origin} \vspace{-4pt}
}
\end{equation}
where $U_{\text{att}}(\mathbf{g}_{\text{rel}}(t)) \in \mathbb{R}$ is the attractive potential at $\mathbf{p}^{\text{IF}}(t)$, which is $\| \mathbf{g}_{\text{rel}}(t) \|^2 /2$. Next, the repulsive force is computed from the repulsive potential $U_{\text{rep}}(\mathcal{L}(t)) \in \mathbb{R}$, where $\mathcal{L}(t) = \{ \mathbf{l}_{1}(t), \mathbf{l}_{2}(t), \ldots, \mathbf{l}_{M}(t) \}$ is the set of $k$-th sensor ray vectors $\mathbf{l}_{k}$ with a resolution of $\tau$ as $M = \lfloor \frac{2\pi}{\tau} \rfloor$. Since we are interested in the ray vectors that detect obstacles (i.e., less than $d_{\text{max}}$), $\mathcal{H} = \{\forall h \in \{1, 2, ..., M\} \mid 0 < \| \mathbf{l}_{h}(t) \| < d_{\text{max}} \}$ includes the indices of such the ray vectors. The repulsive force $\mathbf{F}_{\text{rep}}(L(t)) \in \mathbb{R}^2$ is computed as \vspace{-4pt}
\begin{equation}
\small{
     \mathbf{F}_{\text{rep}}(\mathcal{L}(t)) =  \triangledown U_{\text{rep}}(\mathcal{L}(t)) = 
 \sum_{h \in \mathcal{H}} \frac{-1}{\lVert \mathbf{l}_{h}(t)\rVert^3}\cdot \frac{\mathbf{l}_{h}(t)}{\lVert \mathbf{l}_{h}(t)\rVert}, \label{eq:rep_origin}  \vspace{-4pt}
}
\end{equation}
where $U_{\text{rep}}(\mathcal{L}(t)) \in \mathbb{R}$ is the repulsive potential, which is $\sum^M_{h \in \mathcal{H}}\frac{1}{2 \|  \mathbf{l}_{h}(t) \|^2}$.
The total force vector $\mathbf{F}_{\text{tot}} \in \mathbb{R}^2$ is\vspace{-4pt}
\begin{equation} \label{eq:tot_force_prel}
\small{
    \mathbf{F}_{\text{tot}}(t) = \gamma\mathbf{F}_{\text{att}}(\mathbf{g}_{\text{rel}}(t)) + \sigma\mathbf{F}_{\text{rep}}((\mathcal{L}(t)),   \vspace{-4pt}
}
\end{equation}
where $\gamma\in \mathbb{R}^{+}$ and $\sigma\in \mathbb{R}^{+}$ denote the weight values for the attractive and repulsive forces, respectively.

The total force $\mathbf{F}_{\text{tot}}$, determined by the balance between the weighted attractive and repulsive forces, decides the next moving direction of robots. If the magnitude $\| \mathbf{F}_{\text{tot}} \|$ approaches zero, the robot falls into a local minimum~\cite{yun1997APFlocalMin}.

\vspace{-3pt}
\section{Decentralized Mapless Navigation Methods}
\vspace{-3pt}
\label{sec:method}

\begin{figure}[t]
    \captionsetup{skip=0pt}
    \centering
    \includegraphics[width=0.43\textwidth]{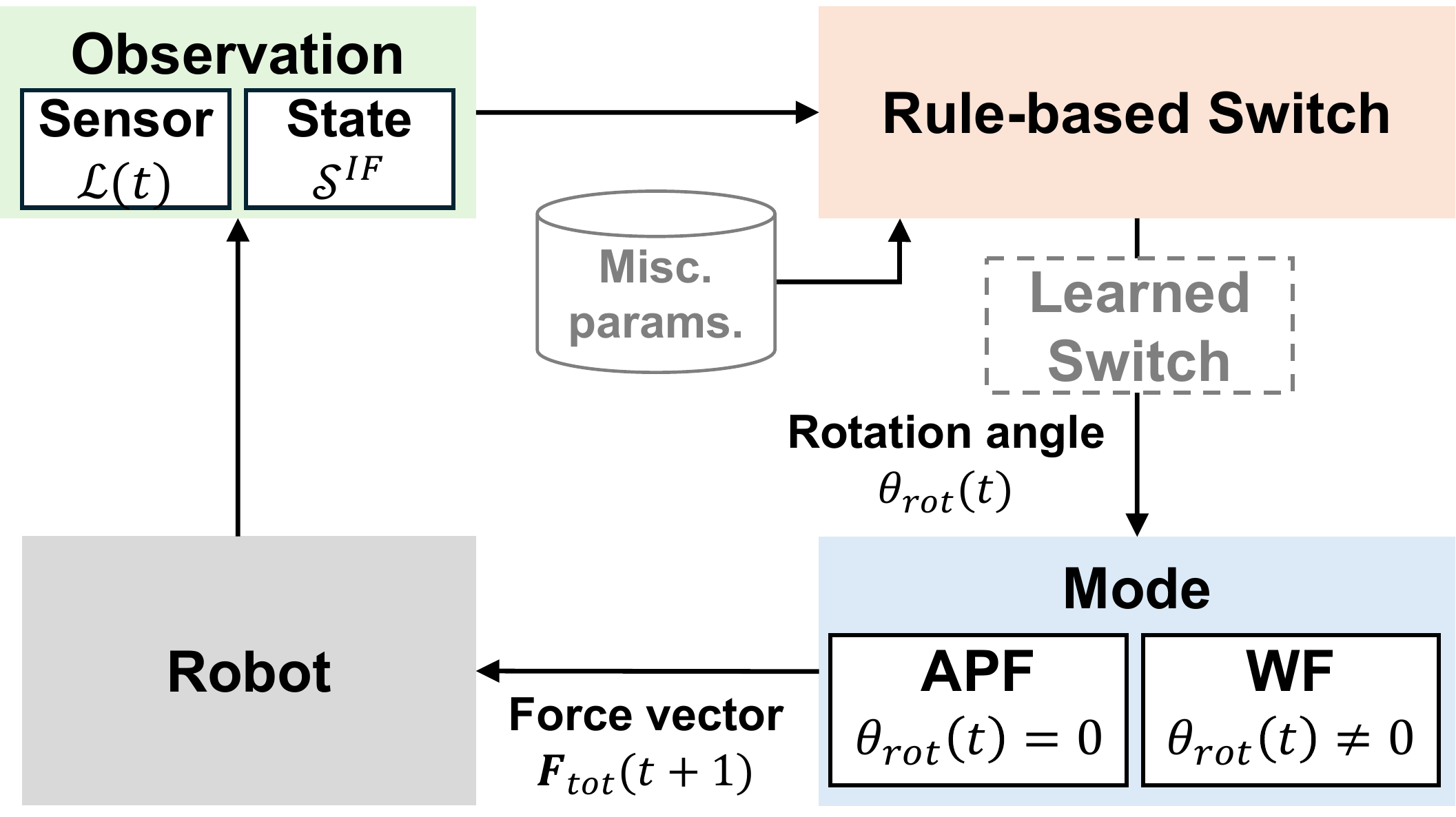}
    \caption{Overview of the proposed system. RS determines the rotation angle $\theta_{\text{rot}}$ based on the observation, controlling the navigation mode. If LS is used together, it overrides the RS output for more efficient navigation learned from human demonstrations.}
    \label{fig:overview} \vspace{-20pt}
\end{figure}

For decentralized mapless navigation, we propose an APF-WF method with two different switching schemes, \textit{Rule-based Switch (RS)} and \textit{Learned Switch (LS)} as abstracted in Fig.~\ref{fig:overview}. Switching between APF and WF depends on the value of $\theta_{\text{rot}}$ produced by RS or LS, where RS uses simple rules and LS overrides the RS output whenever necessary. RS can be used solely where LS is used together with RS. Based on the selection on the mode, the total force vector for the next timestep, $\mathbf{F}_{\text{tot}}(t+1)$, is computed.

In the existing APF-WF methods (e.g., \cite{2023rpf,2008uNAV}), the computed force vector could incur significant motion oscillations owing to successive encounters with dynamic obstacles as their WF solely relies on the tangent direction of the nearest obstacle. To prevent such inefficient navigation, our APF-WF method uses omnidirectional observations to gradually rotate the attractive force, enabling a smoother transition between APF and WF. Specifically, the rotated attractive force $\mathbf{F}'_{\text{att}} = \mathbf{R}(\theta_{\text{rot}}) \cdot \mathbf{F}_{\text{att}}$ alters the next direction of the robot where $\theta_{\text{rot}}$ is the rotation angle and $\mathbf{R} \in \mathbb{R}^{2\times2}$ is a 2D rotation matrix. 
Then $\mathbf{F}'_{\text{att}}$ is normalized to have the maximum magnitude $d_{\text{max}}$, preventing the force from having excessive influence in the direction opposite to the goal:\vspace{-3pt}
\begin{equation}
\small{
    \mathbf{F}'_{\text{att}}(\mathbf{g}_{\text{rel}}) = d_{\text{max}} \frac{\mathbf{g}_{\text{rel}}(t)}{\| \mathbf{g}_{\text{rel}}(t) \|}.
}\vspace{-3pt}
\end{equation} 
The total force is given by\vspace{-3pt}
\begin{equation}
\small{
    \mathbf{F}_{\text{tot}}(t) = \omega\mathbf{F}'_{\text{att}}(\mathbf{g}_{\text{rel}}(t)) + (1-\omega)\mathbf{F}_{\text{rep}}((\mathcal{L}(t)) \label{eq:tot_f_mthd}
}\vspace{-3pt}
\end{equation} 
where $\omega$ $(0 < \omega < 1)$ controls the balance between the attractive and repulsive forces. 

\subsection{Rule-based Switch (RS)} 
\label{sec:RS}
\vspace{-3pt}

RS determines the mode based on $\theta_{\text{rot}}$, which is updated at each $t$: APF if $\theta_{\text{rot}}(t) = 0$ and WF otherwise. Each robot maintains $\mathbf{s}^{\text{IF}}(t)$, $\mathbf{s}^{\text{IF}}(t_{\text{hp}})$, and $\mathbf{s}^{\text{IF}}(t_{\text{lp}})$, which are states at $t$, $t_{\text{hp}}$, and $t_{\text{lp}}$, respectively. Here, $t_{\text{hp}}$ and $t_{\text{lp}}$ are the times at \textit{Hit Point (HP)} and \textit{Leave Point (LP)}, respectively. HP indicates the transition from APF to WF while LP represents the reverse switch~\cite{2008uNAV}. Along with the states at $t$, the same states at $t-1$ regarding the previous HP and LP are stored in $\mathcal{S}^{\text{IF}}$. These information help prevent the robot from repeating looped paths around obstacles.

At every $t$, $\mathbf{s}^{\text{IF}}(t)$ is updated while $\mathbf{s}^{\text{IF}}(t_{\text{lp}})$ is updated if the transition occurs. On the other hand, $\mathbf{s}^{\text{IF}}(t_{\text{hp}})$ is updated only if the transition occurs and the distance from the current position to the goal, $\| \mathbf{g}_{\text{rel}}(t) \|$, is smaller than the distance from the previously stored HP to the goal, $\| \mathbf{g}_{\text{rel}}(t_{\text{hp}}) \|$. This constraint on updating HP prevents robots from moving farther from the goal.

\begin{algorithm}[t]
\small
\caption{\textsc{Rule-based Switch (RS)}}
\label{alg1:RS}
\begin{algorithmic}[1]
\renewcommand{\algorithmicrequire}{\textbf{Input:} }
\renewcommand{\algorithmicensure}{\textbf{Output: }}

\REQUIRE $\mathbf{g}^{\text{IF}}, \mathcal{S}^{\text{IF}}, \mathcal{L}(t), \theta_{\text{rot}}(t-1), I_{\text{dir}}(t-1), I_{\text{dir}}(t_{\text{hp}}), f_{\text{thr}},$ \\ $ \qquad  \theta_{\text{upd}}, \theta_{\text{rcv}}, M$ \label{alg1:RS_input}  
\ENSURE $\theta_{\text{rot}}(t), I_{\text{dir}}(t)$ \label{alg1:RS_output}

\STATE {$\mathbf{g}_{\text{rel}}(t) \gets \textsc{RelativeGoal}(\mathcal{S}^{\text{IF}}(t), \mathbf{g}^{\text{IF}})$} \label{alg1:relativeGoal}

\STATE {$\mathbf{F}_{\text{tot}}(t), \mathbf{F}_{\text{att}}(t), \mathbf{F}_{\text{rep}}(t)  \gets \textsc{GetForces}(\mathcal{L}(t), \mathbf{g}_{\text{rel}}(t))$}  \label{alg1:line2}

\STATE {$revisitHitPoint \gets \textsc{CheckLoop}(\mathcal{S}^{\text{IF}})$} \label{alg1:checkloop}

\IF {$revisitHitPoint$} \label{alg1:line4}
    \STATE {$I_{\text{dir}}(t) \gets -I_{\text{dir}}(t_{\text{hp}})$} \label{alg1:opposite_dir}
\ELSIF {$\theta_{\text{rot}}(t-1) \neq 0$} \label{alg1:line6}
    \STATE{$I_{\text{dir}}(t) \gets I_{\text{dir}}(t-1)$} \label{alg1:keep_dir}
\ELSE \label{alg1:line8}
    \STATE {$I_{\text{dir}}(t) \gets \textsc{ChooseDir}(\mathcal{L}(t), \mathcal{S}^{\text{IF}}(t), \mathbf{g}_{\text{rel}}(t), M$}) \label{alg1:choosedir} 
\ENDIF \label{alg1:line10}

\IF { $\| \mathbf{F}_{\text{tot}}(t) \| < f_{\text{thr}}$} \label{alg1:line11} 

\STATE{$\theta_{\text{rot}}(t) \gets \theta_{\text{rot}}(t-1)  + I_{\text{dir}}(t)\cdot \theta_{\text{upd}}$ } \label{alg1:line12}  

\ELSE \label{alg1:line13}

\STATE{$\theta_{\text{rot}}(t) \gets \theta_{\text{rot}}(t-1)  - I_{\text{dir}}(t)\cdot \theta_{\text{rcv}}$ } \label{alg1:line14}

\IF {$(I_{\text{dir}}(t)\cdot \theta_{\text{rot}}(t) < 0)$} \label{alg1:line15}
        \STATE {$\theta_{\text{rot}}(t) \gets 0$} \label{alg1:line16}
\ENDIF \label{alg1:line17}
\ENDIF \label{alg1:line18}

\IF {$\textsc{BreakWF}(\mathcal{S}^{\text{IF}}(t)$) } \label{alg1:breakWF_start} \label{alg1:line19}
\STATE $\theta_{\text{rot}}(t) \gets 0$ \label{alg1:line20} 
\ENDIF \label{alg1:breakWF_end} \label{alg1:line21}


\RETURN {$\theta_{\text{rot}}(t)$, $I_{\text{dir}}(t)$}  \label{alg2:line22}

\end{algorithmic}
\end{algorithm}

In Alg.~\ref{alg1:RS}, the goal $\mathbf{g}^{\text{IF}}$ is transformed into the relative goal using \textsc{RelativeGoal} (line~\ref{alg1:relativeGoal}), which is then used by \textsc{GetForces} to compute force vectors by (\ref{eq:att_origin}--\ref{eq:rep_origin}) and (\ref{eq:tot_f_mthd}) to check the condition for switching.
\textsc{CheckLoop} in line~\ref{alg1:checkloop} returns true if the robot repeats a loop in its path. This function can be implemented in various ways; in our case, we use the historical records $\mathcal{S}^{\text{IF}}$ to check if the robot revisits a previous HP after leaving it. In the presence of noisy sensor or control, a loop may be detected after repeated revisits to account for possible accidental revisits to the previous HP.

The value of $I_{\text{dir}} \in \{-1, 1\}$ determines the direction for WF: $I_{\text{dir}}=-1$ makes the robot follow the wall in the clockwise direction, while $I_{\text{dir}}=1$ causes it to follow the wall in counterclockwise. If the robot has revisited the previous HP, the WF direction is reversed from the direction chosen at the HP not to repeat the same behavior (line~\ref{alg1:opposite_dir}). If the robot is already in WF mode (i.e., $\theta_{\text{rot}}(t-1) = 0$), the direction remains unchanged (line~\ref{alg1:keep_dir}). If the robot is not in a looped path and it is converting to WF, the WF direction is computed by \textsc{ChooseDir}  (line~\ref{alg1:choosedir}, detailed in Alg.~\ref{alg2:choosedir}), which selects either of the WF directions that brings the robot closer to the goal than the other direction. It first calculates $\phi_{\text{min}}$, the orientation of the sensor ray $\mathbf{l}_{k}(t)$ with the minimum distance to goal $\mathbf{g}_{\text{rel}}(t)$ (Alg.~\ref{alg2:choosedir} lines~\ref{alg2:get_j}--\ref{alg2:line2}). After an angle wrapping (to make it within $(-\pi, \pi]$), Alg.~\ref{alg2:choosedir} selects the direction that requires the least amount of rotation (Alg.~\ref{alg2:choosedir} lines~\ref{alg2:line8}--\ref{alg2:line14}).

\begin{algorithm}[t]
\small
\caption{\textsc{ChooseDir}}
\label{alg2:choosedir}
\begin{algorithmic}[1]
\renewcommand{\algorithmicrequire}{\textbf{Input:} }
\renewcommand{\algorithmicensure}{\textbf{Output: }}

\REQUIRE $\mathcal{L}(t), \mathcal{S}^{\text{IF}}(t), \mathbf{g}_{\text{rel}}(t), M$
\ENSURE $I_{\text{dir}}$

\STATE {$j \gets \argmin_{k \in \{1, 2, \ldots, M\}} (\| \mathbf{g}_{\text{rel}}(t)) - \mathbf{l}_{k}(t) \|)$} \label{alg2:get_j}
\STATE {$\phi_{\text{min}} \gets \frac{2\pi (j-1)}{M}$} \label{alg2:line2}
    
\IF  {$\phi_{\text{min}} > 2\pi$} \label{alg2:line3}
    \STATE {$\phi_{\text{min}} \gets \phi_{\text{min}} - 2\pi$} \label{alg2:line4}
\ELSIF{$\theta_{\text{min}} \leq -2\pi$} \label{alg2:line5}
    \STATE {$\phi_{\text{min}} \gets \phi_{\text{min}} + 2\pi$} \label{alg2:line6}
\ENDIF \label{alg2:line7}

\STATE {$\phi_{\text{goal}}(t) \gets \text{atan2}(y_{\text{rel}}(t),x_{\text{rel}}(t))$} \label{alg2:line8}
\STATE {$\phi_{\text{dir}} \gets \phi_{\text{min}} - \phi_{\text{goal}}(t)$} \label{alg2:line9}
    
\IF {$ 0 < \phi_{\text{dir}} < \pi$ \OR $ -2\pi<\phi_{\text{dir}}<-\pi$} \label{alg2:line10}
    \STATE {$I_{\text{dir}} \gets 1$} \label{alg2:line11}
\ELSIF{$ -\pi < \phi_{\text{dir}} < 0$ \OR $ \pi < \phi_{\text{dir}} < 2\pi$} \label{alg2:line12}
    \STATE {$I_{\text{dir}} \gets -1$} \label{alg2:line13}

\ENDIF \label{alg2:line14}

\RETURN {$I_{\text{dir}}$} \label{alg2:line15}
\end{algorithmic}
\end{algorithm}

Based on the total force vector and WF direction, Alg.~\ref{alg1:RS} updates $\theta_{\text{rot}}(t)$ for the robot to continuously rotate the direction of the attractive force to escape from local minima using WF and then return to APF. If the robot approaches to a local minimum where the robot looses the force to move, the magnitude of $\mathbf{F}_{\text{tot}}(t)$ becomes small. If the magnitude goes below $f_{\text{thr}}$ (a threshold value experimentally set to $0.5 \cdot  d_{\text{max}}$), the robot updates $\theta_{\text{rot}}(t)$ by adding $\theta_{\text{upd}} \in \mathbb{R}^{+}$ multiplied by the WF direction $I_{\text{dir}}$ (line~\ref{alg1:line12}). A small value of $\theta_{\text{upd}}$ allows the robot to gradually change $\mathbf{F_{\text{att}}}$, ensuring smooth directional adjustments. If $ \| \mathbf{F}_{\text{tot}}(t) \| $ is greater than or equal to $f_{\text{thr}}$, $\theta_{\text{rot}}(t)$ decreases by subtracting a recovery angle $\theta_{\text{rcv}}$, multiplied by $I_{\text{dir}}$, allowing the robot to gradually return to APF mode by reducing the rotation angle $\theta_{\text{rot}}$ to zero once it has cleared the local minimum (lines~\ref{alg1:line14}--\ref{alg1:line16}). In our implementation, we experimentally set the values of $\theta_{\text{upd}}$ and $\theta_{\text{rcv}}$ to be $\frac{2\pi}{M}$ and $0.5\cdot\theta_{\text{upd}}$, respectively. 

The updated $\theta_{\text{rot}}(t)$ adjusts the direction of $\mathbf{F}'_{\text{att}}$, guiding the total force to steer the robot away from the local minimum. Even if the obstacle is large and nonconvex, which is likely to incur a local minimum, the robot will continue to rotate the attractive force incrementally and move along the obstacle boundary, with its fixed WF direction determined by line~\ref{alg1:keep_dir}. Finally, the robot can escape from the local minimum, at which point the rotation angle is reset to zero, and then stops WF to move toward the goal (lines~\ref{alg1:line20}). This decision is made by \textsc{BreakWF} which is inspired by the M-Line condition presented in Bug2~\cite{lumelsky1986bug1_bug2}. \textsc{BreakWF} returns true if $\mathbf{g}^{\text{IF}}_{\text{rel}}(t)$ aligns with $\mathbf{g}^{\text{IF}}_{\text{rel}}(t_{\text{hp}})$\footnote{The alignment means that the robot is on a position such that the line segment between the position and the goal is the shortest.}, and the robot has moved closer to the goal than it was at HP. This switching mechanism minimizes unnecessary detours, allowing for a more efficient exit from WF as the robot approaches the goal. 

\subsection{Learned Switch (LS)} 
\label{sec:LS}
\vspace{-3pt}

While RS performs effectively in many scenarios, its simple switching rule can struggle with complex multi-robot interactions, leading to false-positive WF behaviors, referred to as \textit{false-WF}. Two key cases illustrate this issue. First, robots may mistakenly follow each other, treating one another as static obstacles (see Fig.~\ref{fig:1_b} left), causing them to endlessly WF until approaching other obstacles. Second, in narrow passages, when robots face each other, both may start WF simultaneously and move backward, though a more efficient solution would have one robot move forward while the other steps back to pass the passage in one direction. 

To address this issue, we propose LS, an extension of RS that incorporates a Neural Network (NN)-based switch using a Vision Transformer (ViT) encoder~\cite{dosovitskiy2020vit}. LS is trained on human demonstration data where an expert provides switching decisions between APF and WF. The input to LS is an observation vector $\mathbf{o}_{\text{obs}}(t)$ of size $M+17$ with $M$ range values and $17$ dimensions of relevant states at $t$: (i) the 2D relative positions to the goal from the current and previous states, the latest HP and LP ($8$ dimensions),  (ii) the current heading of the robot ${\psi}(t)$, (iii) the $4$-dim rotation angles from the current and previous states, the latest HP and LP, (iv) the range value of the ray $z$, $\| \mathbf{l}_z(t) \|$, where $z$ the index of the ray closest to the goal, (v) the magnitude $\| \mathbf{F}_{\text{tot}}(t) \|$, (vi) the WF direction at HP $I_{\text{dir}}(t_{\text{hp}})\in \mathbb{R}$, and (vii) the indicator variable $I_{\text{mode}}(t)\in \{0, 1\}$ (i.e., APF or WF, respectively).
The information of $M$ sensor rays and (i)--(iv) allow LS to assess both the surrounding obstacles and the past trajectory of the robot to extract features to learn from human decision-making patterns which can be implied by the values (v)--(vii).

These vectors are stacked into an observation matrix $\textbf{O}(t) \in \mathbb{R}^{T{\text{seq}} \times (M+17)}$ over $T_{\text{seq}}$ timesteps. For  $t < T_{\text{seq}}$, the matrix is padded by repeating the initial observation vector $\mathbf{o}_{\text{obs}}(0)$. The encoder architecture is with a patch size of $1 \times (M+17)$, allowing the model to apply Multi-Head Attention across the timesteps. After the transformer encoder layers, a Multi-Layer Perceptron (MLP) classifier with $3$ fully connected layers and ReLU activation is attached. The final LS architecture consists of $3$ layers of transformer encoders for feature extraction, followed by $3$ fully connected layers for classification, with both the embedding and MLP dimensions set to $512$. LS evaluates the sequential observations to decide whether to override the current RS decision. LS outputs either $0$ (APF mode) or $1$ (WF mode).

\begin{figure}[t]
    \captionsetup{skip=0pt}
    \centering
    \includegraphics[width=0.35\textwidth]{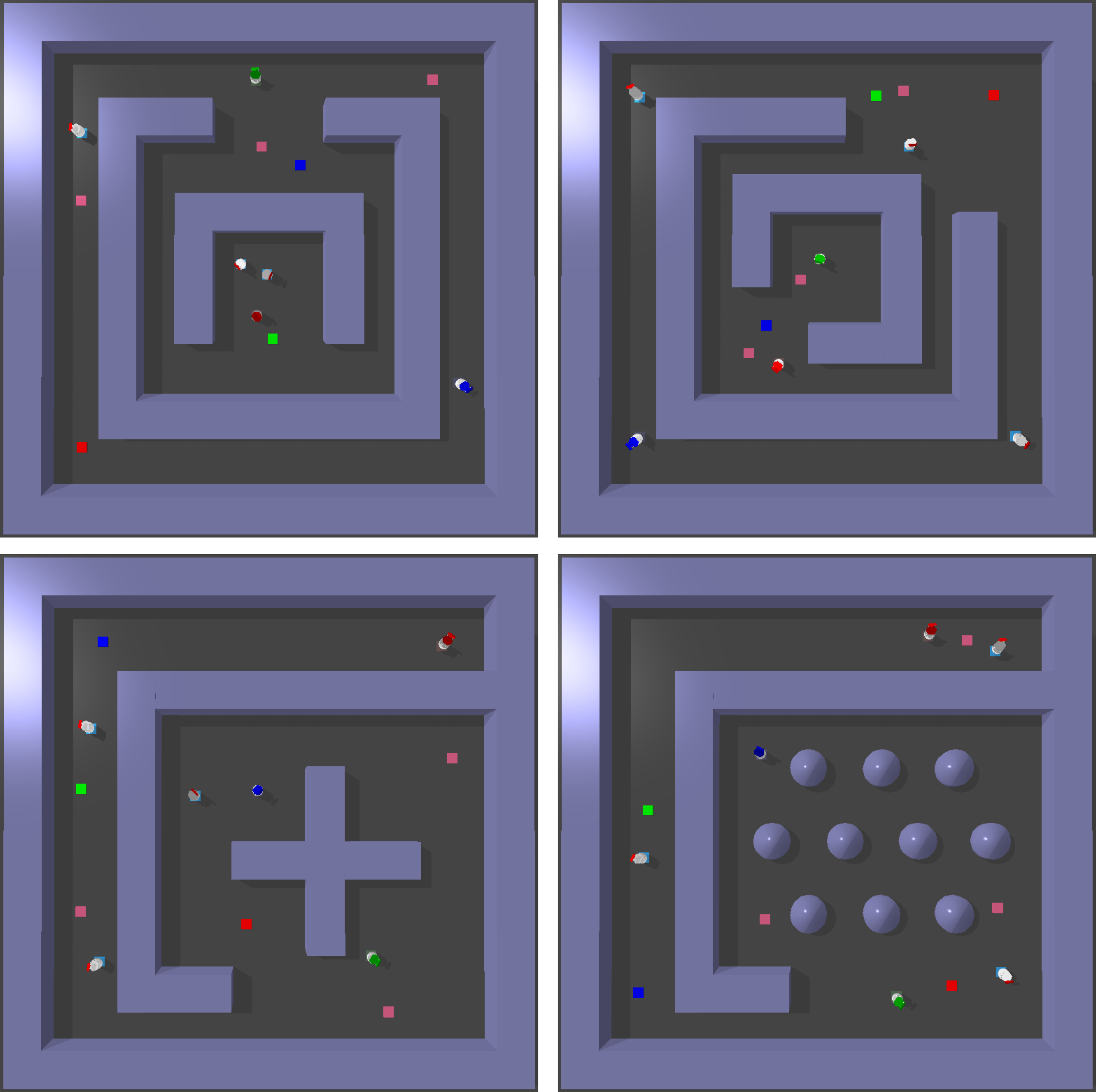}
    \caption{Four environments for data collection. Three colored robots can be intervened by a human expert while white robots perform as dynamic obstacles. Initial robot positions and goals are randomly generated.}
    \label{fig:TrainMap} \vspace{-22pt}
\end{figure}

The data are collected in four environments in simulation shown in Fig.~\ref{fig:TrainMap}. They are designed to cause various difficult situations incurred by nonconvex obstacles and robot encounters. A human expert observes six robots from the top view. The expert is allowed to toggle the mode of three robots (red, green, and blue) using keyboard inputs (e.g., pressing 1, 2, and 3 for each robot). The rest three robots are considered dynamic obstacles. We collect $1800$ demonstration data points, with $450$ data points per environment, resulting in approximately $1.2$ million timesteps in total. For binary classification, we employed Binary Cross Entropy Loss (BCELoss)~\cite{zhang2018bceloss}. The dataset is split into training and testing sets with an $80:20$ ratio. Training is conducted over $50$ epochs with a learning rate of $3\times10^{-4}$ using the Adam optimizer~\cite{diederik2014adam}. The model with the highest classification accuracy on the test set is selected for final evaluation.

\section{Experiments}
\label{sec:exp}
\vspace{-3pt}

\begin{figure}[t]
\vspace{-2pt}
\captionsetup{skip = 0pt}
    \centering
    \begin{subfigure}[b]{\columnwidth}
        \centering
        \captionsetup{skip = 0pt}
        \includegraphics[width=\textwidth]{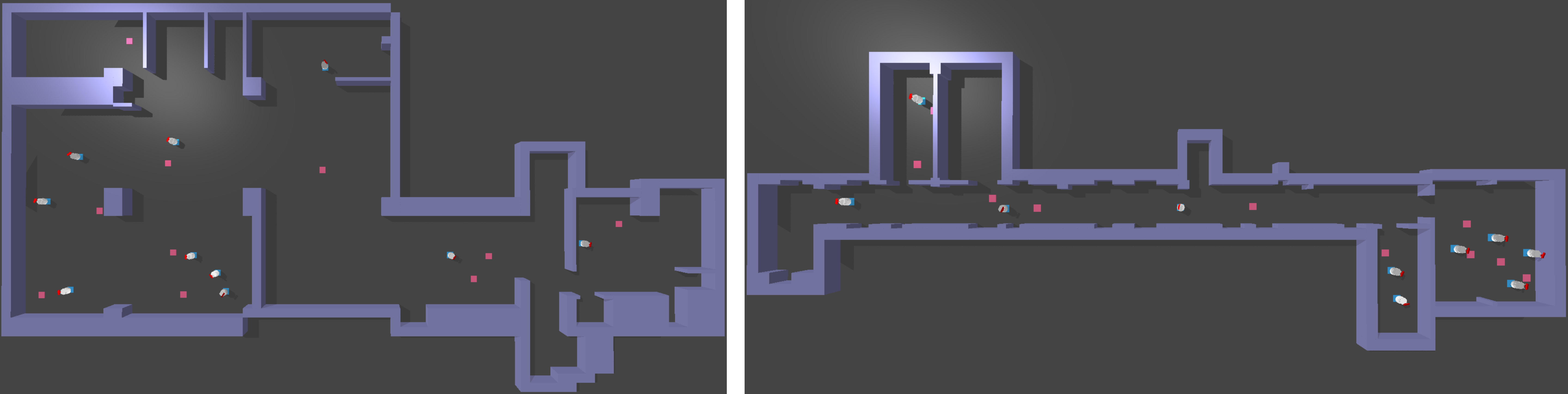}
        \caption{Real-world layouts}
        \label{fig:realistic_env}
    \end{subfigure}
    \hfill 
    \begin{subfigure}[b]{\columnwidth}
        \centering
        \captionsetup{skip = 0pt}
        \includegraphics[width=\textwidth]{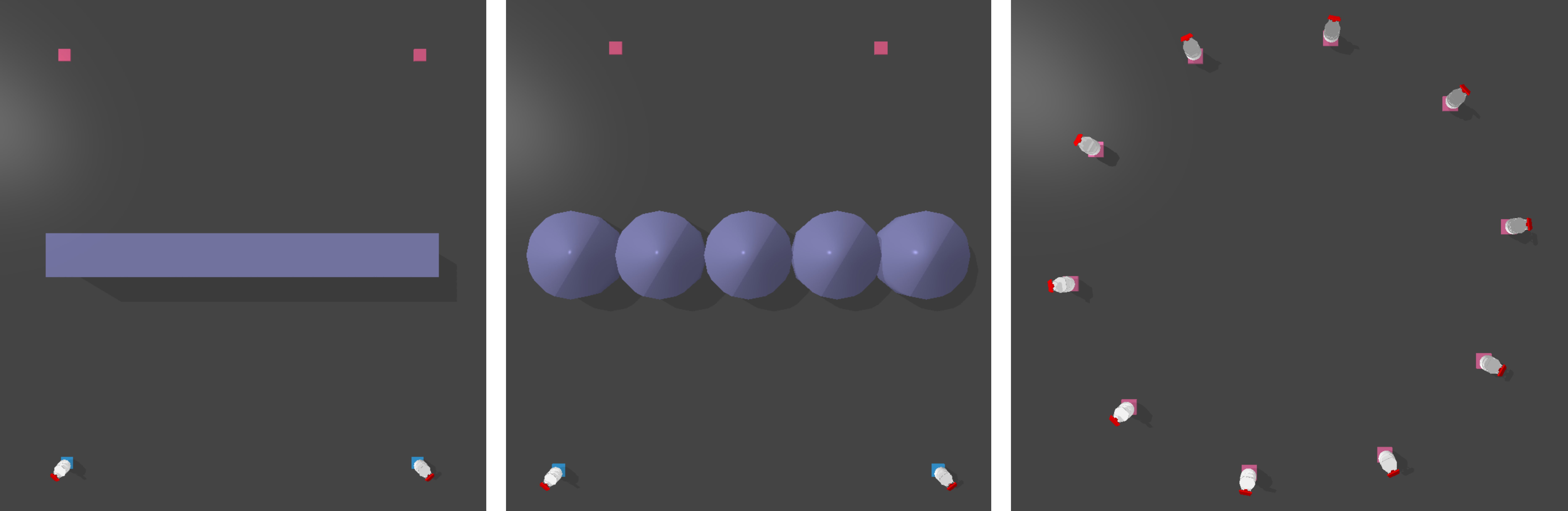}
        \caption{Symmetric layouts}
        \label{fig:symmetric_env}
    \end{subfigure}
    \caption{Test environments where robots are in white, goals in pink, and static obstacles in purple. (a) Real-world layouts with \textsf{Nakwon} (left) and \textsf{Sogang} (right) where robots and goals are placed at random. (b) Symmetric layouts with \textsf{Flat}, \textsf{Cylind}, and \textsf{Swap} from the left.} 
    \label{fig:TestEnvs}\vspace{-20pt}
\end{figure}

\subsection{Experimental Setup}
\vspace{-3pt}
\subsubsection{Environments} 
We use two sets of test environments. \textit{Real-world layouts} include the nonconvex environments whose floor plans are from real buildings. In Fig.~\ref{fig:realistic_env}, \textsf{Nakwon} (left) has few bottleneck areas can cause congestion or even deadlocks. Up to $30$ robots are randomly generated in the test instances. \textsf{Sogang} in the right is more challenging as it is with a long narrow corridor, small rooms, and doorways, which incur frequent encounters and conflicts among robots. Since the entire space is small, up to $10$ robots are used. 
\textit{Symmetric layouts} consist of synthetic environments which are likely to incur difficult situations. In \textsf{Flat} (Fig.~\ref{fig:symmetric_env} left), the robots often recognize each other as static obstacles if an ordinary WF as described in Fig.~\ref{fig:1_b}. \textsf{Cylind} (Fig.~\ref{fig:symmetric_env} mid) is even more challenging with the concave spots between the cylindrical pillars. \textsf{Swap} causes significant congestion in the center (Fig.~\ref{fig:symmetric_env}, right), as up to $10$ robots attempt to exchange positions across a $10$\,m distance. Although \textsf{Swap} does not have any static obstacle, deadlocks and the false-WF are common due to symmetry. In each environment, $20$ random instances (real-world) and $300$ instances (symmetric) are generated for each team size.

\subsubsection{Baselines} We compare our APF-RS and APF-LS with four baseline methods. ORCA~\cite{van2008orca} and vanilla APF~\cite{apf} are commonly prone to deadlocks and local minima. $\mu$NAV~\cite{2008uNAV}, and Reinforced Potential Field (RPF)~\cite{2023rpf} are compared as the representative APF-WF methods.

\subsubsection{Metrics} Four metrics are used: (i) \textit{success rate} where an instance is successful if \textit{all} robots reach their goals  within $T$ without collisions, (ii) \textit{arrival rate} measuring the proportion of robots reaching their goals within $T$ without collisions, (iii) \textit{makespan} referring to the maximum timesteps until a success\footnote{In failed instances, makespan is not available.}, (iv) \textit{mean timestep} averaging the number of timesteps taken to reach the goals without collisions.

\subsubsection{Others} We use TurtleBot4 in both simulation and physical experiments where $r = 0.17$\,m, which is equipped with a ray sensor where $d_{\text{max}} = 10$\,m and $M = 100$. Simulations are done in PyBullet~\cite{coumans2016pybullet} with 5\,Hz control frequency. The system is with an AMD Ryzen 5900X CPU, NVIDIA GeForce RTX 3070 Ti GPU, and 32GB RAM.

\subsection{Tests in Simulation} 
\vspace{-3pt}

\subsubsection{Real-world layouts} 
\begin{figure}
\vspace{-5pt}
\captionsetup{skip = 0pt}
    \centering
    \ \ \begin{subfigure}[b]{0.99\columnwidth}
        \centering
        \captionsetup{skip = -2pt}
        \includegraphics[width=\textwidth]{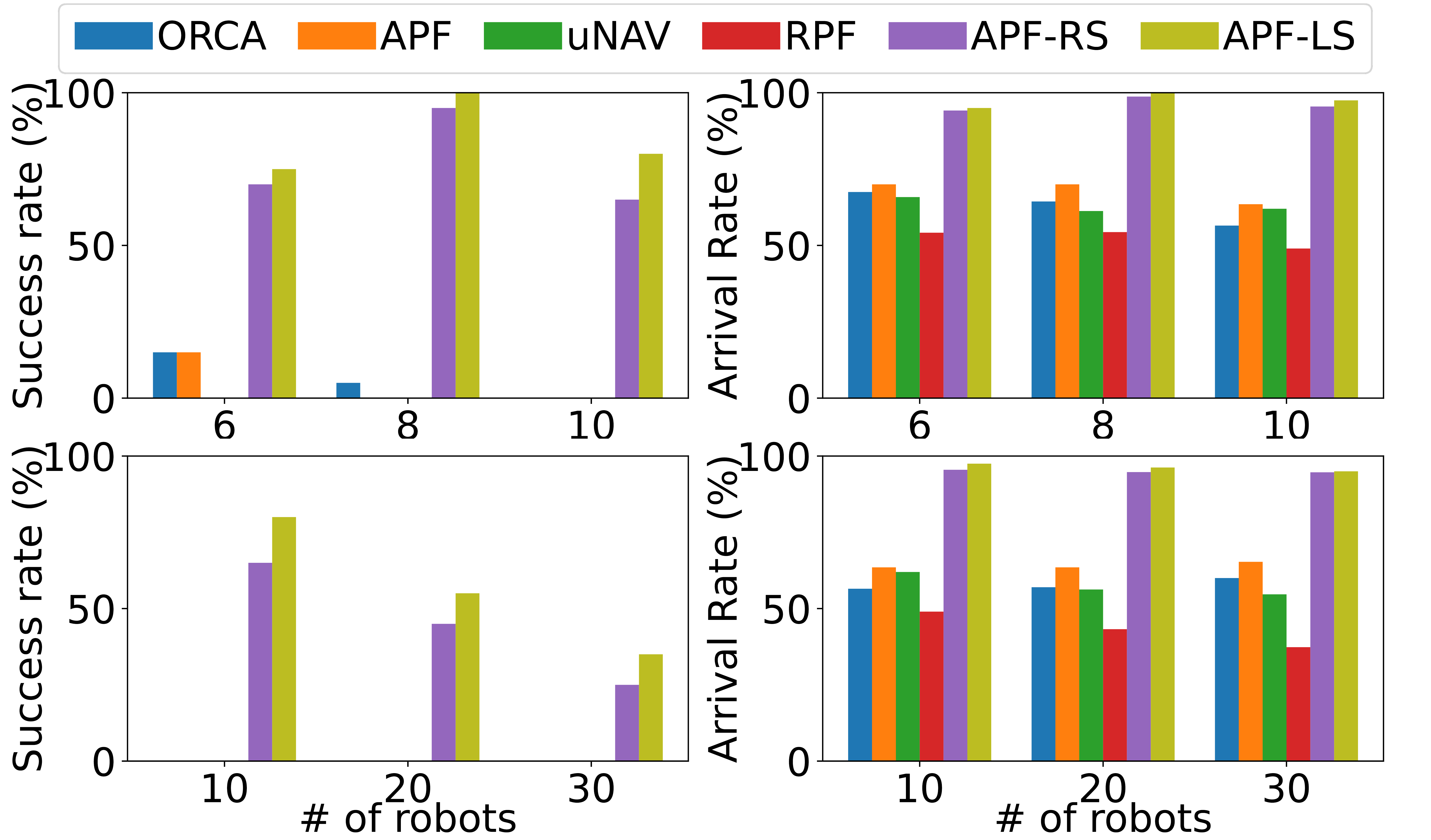}
        \caption{Result of \textsf{Nakwon}}
        \label{fig:nakwon}
    \end{subfigure}
    \begin{subfigure}[b]{0.96\columnwidth}
        \centering
        \captionsetup{skip = -2pt}
        \includegraphics[width=\textwidth]{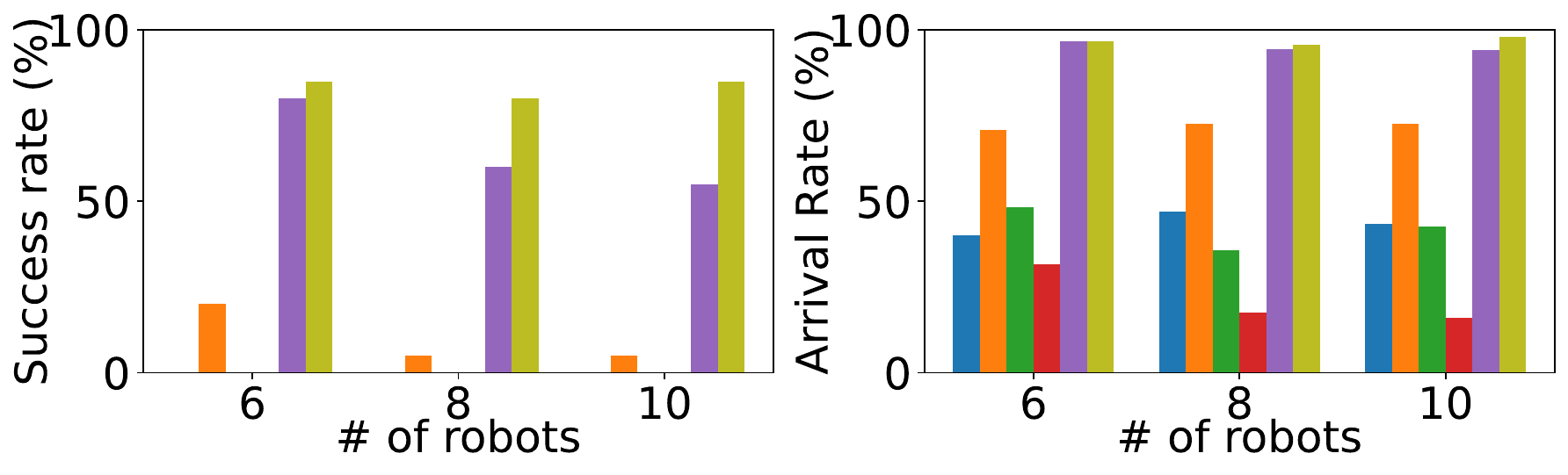}
        \caption{Result of \textsf{Sogang}}
        \label{fig:sogang}
    \end{subfigure}
    \caption{Success and arrival rates of real-world layouts} \vspace{-22pt}
    \label{fig:Real_test_results}
\end{figure}

The success and arrival rates are summarized in Fig.~\ref{fig:Real_test_results} where makespan and mean timestep are shown in Table~\ref{tab:real_table}. In this set, we have $T = 3000$ for \textsf{Sogang} and $4000$ for \textsf{Nakwon}. In \textsf{Nakwon}, the success rates of our methods range from $60\%$ to $100\%$ in tests with $6$ to $10$ robots, whereas the baselines achieves very low success rates, below $15\%$. As the number of robots increases, our methods also show declines but neither of the baselines succeed. The arrival rates of RS and LS are at least $94.2\%$, with improvements of $24.2\%$ to $34\%$ over the baselines, respectively. Similar results are observed in \textsf{Sogang}. 

A noteworthy observation is the lower success rates of RS and LS with $N=6$ compared to $N=8$, despite $N=6$ being an ostensibly easier case. In \textsf{Nakwon}, which has large open spaces and few bottleneck areas, more robots tend to follow long walls for extended periods. With $8$ robots, more frequent encounters between robots promote interactions rather than prolonged wall-following, leading to quicker goal attraction. However, this positive effect diminishes with $10$ robots, as increased inter-robot interactions cause conflicts, negatively impacting navigation performance. Despite this, our methods consistently outperform the baselines for all robot counts. Excluding APF, all other compared methods fail completely, while our approach improves success rates by up to 65\% over APF.

Although \textsf{Sogang} is more confined, LS trained on human demonstrations consistently maintains high success rates above $80\%$, regardless of the number of robots. In contrast, RS shows a gradual decline, indicating that learning from human decisions can effectively mitigate inter-robot conflicts. It is important to clarify that the higher makespan and mean timestep for APF-RS and APF-LS compared to the baselines do not necessarily imply lower performance. These metrics are only calculated for successful instances where data points from the baselines are quite scarce (up to four). The baselines succeed in relatively easy instances, naturally resulting in lower values for these metrics. Conversely, our methods succeed in a broader range of instances, including more difficult ones, which increases the values.

Overall, our methods outperform the baselines in constrained real-world environments where neither maps nor inter-robot communication are available.

\begin{table*}[t]
\captionsetup{skip=0pt}
\caption{Makespan and mean timestep of real-world layouts (standard deviations in the parentheses)}
\label{tab:real_table}
\centering
\resizebox{2.0\columnwidth}{!}{%
\begin{tabular}{|c|c|cccccccccccc|}
\hline
\multirow{2}{*}{Env.} &
  \multirow{2}{*}{\#robot}
   &
  \multicolumn{6}{c|}{Makespan} &
  \multicolumn{6}{c|}{Mean timestep} \\ \cline{3-14} 
 &
   &
  \multicolumn{1}{c|}{ORCA} &
  \multicolumn{1}{c|}{APF} &
  \multicolumn{1}{c|}{$\mu$NAV} &
  \multicolumn{1}{c|}{RPF} &
  \multicolumn{1}{c|}{APF-RS} &
  \multicolumn{1}{c|}{APF-LS} &
  \multicolumn{1}{c|}{OCRA} &
  \multicolumn{1}{c|}{APF} &
  \multicolumn{1}{c|}{$\mu$NAV} &
  \multicolumn{1}{c|}{RPF} &
  \multicolumn{1}{c|}{APF-RS} &
  APF-LS \\ \hline

\multirow{5}{*}{\textsf{Nakwon}} &
  6 &
  \multicolumn{1}{c|}{297.7 (155.7)} &
  \multicolumn{1}{c|}{247.7 (70.2)} &
  \multicolumn{1}{c|}{-} &
  \multicolumn{1}{c|}{-} &
  \multicolumn{1}{c|}{1465.5 (1551.5)} &
  \multicolumn{1}{c|}{1620.5 (1602.5)} &
  \multicolumn{1}{c|}{164.4 (40.8)} &
  \multicolumn{1}{c|}{1135.4 (698.3)} &
  \multicolumn{1}{c|}{459.6 (183.0)} &
  \multicolumn{1}{c|}{213.3 (124.5)} &
  \multicolumn{1}{c|}{729.7 (591.5)} &
  702.2(529.2) \\ \cline{2-14} 
 &
  8 &
  \multicolumn{1}{c|}{560.0 (0.0)} &
  \multicolumn{1}{c|}{-} &
  \multicolumn{1}{c|}{-} &
  \multicolumn{1}{c|}{-} &
  \multicolumn{1}{c|}{2668.3 (1198.6)} &
  \multicolumn{1}{c|}{2533.4 (1249.8)} &
  \multicolumn{1}{c|}{169.6 (55.9)} &
  \multicolumn{1}{c|}{1195.4 (542.5)} &
  \multicolumn{1}{c|}{516.3 (256.6)} &
  \multicolumn{1}{c|}{201.1 (53.7)} &
  \multicolumn{1}{c|}{781.7 (477.1)} &
  742.7(480.2) \\ \cline{2-14} 
 &
  10 &
  \multicolumn{1}{c|}{-} &
  \multicolumn{1}{c|}{-} &
  \multicolumn{1}{c|}{-} &
  \multicolumn{1}{c|}{-} &
  \multicolumn{1}{c|}{2928.8 (1311.7)} &
  \multicolumn{1}{c|}{2990.8 (1150.4)} &
  \multicolumn{1}{c|}{157.3 (39.3)} &
  \multicolumn{1}{c|}{1415.1 (543.4)} &
  \multicolumn{1}{c|}{444.6 (164.8)} &
  \multicolumn{1}{c|}{214.5 (63.5)} &
  \multicolumn{1}{c|}{855.6 (414.6)} &
  752.8(309.8) \\ \cline{2-14} 
 &
  20 &
  \multicolumn{1}{c|}{-} &
  \multicolumn{1}{c|}{-} &
  \multicolumn{1}{c|}{-} &
  \multicolumn{1}{c|}{-} &
  \multicolumn{1}{c|}{3434.7 (507.2)} &
  \multicolumn{1}{c|}{3487.8 (570.1)} &
  \multicolumn{1}{c|}{150.1 (31.9)} &
  \multicolumn{1}{c|}{1537.7 (282.9)} &
  \multicolumn{1}{c|}{555.6 (139.4)} &
  \multicolumn{1}{c|}{288.2 (131.1)} &
  \multicolumn{1}{c|}{906.8 (238.8)} &
  828.4(273.3) \\ \cline{2-14} 
 &
  30 &
  \multicolumn{1}{c|}{-} &
  \multicolumn{1}{c|}{-} &
  \multicolumn{1}{c|}{-} &
  \multicolumn{1}{c|}{-} &
  \multicolumn{1}{c|}{3896.0 (57.9)} &
  \multicolumn{1}{c|}{3737.4 (232.6)} &
  \multicolumn{1}{c|}{163.6 (34.5)} &
  \multicolumn{1}{c|}{1447.3 (370.9)} &
  \multicolumn{1}{c|}{632.5 (100.4)} &
  \multicolumn{1}{c|}{349.4 (153.3)} &
  \multicolumn{1}{c|}{954.7 (149.1)} &
  843.1(186.8) \\ \hline
\multirow{3}{*}{\textsf{Sogang}} &
  6 &
  \multicolumn{1}{c|}{-} &
  \multicolumn{1}{c|}{609.2 (72.9)} &
  \multicolumn{1}{c|}{-} &
  \multicolumn{1}{c|}{-} &
  \multicolumn{1}{c|}{1749.1 (850.8)} &
  \multicolumn{1}{c|}{1822.6 (802.4)} &
  \multicolumn{1}{c|}{223.8 (102.2)} &
  \multicolumn{1}{c|}{910.5 (538.9)} &
  \multicolumn{1}{c|}{589.1 (329.2)} &
  \multicolumn{1}{c|}{206.8 (165.2)} &
  \multicolumn{1}{c|}{653.6 (308.0)} &
  649.1(302.2) \\ \cline{2-14} 
 &
  8 &
  \multicolumn{1}{c|}{-} &
  \multicolumn{1}{c|}{609.0 (0.0)} &
  \multicolumn{1}{c|}{-} &
  \multicolumn{1}{c|}{-} &
  \multicolumn{1}{c|}{2058.8 (868.9)} &
  \multicolumn{1}{c|}{2119.7 (842.4)} &
  \multicolumn{1}{c|}{313.6 (151.4)} &
  \multicolumn{1}{c|}{967.2 (378.3)} &
  \multicolumn{1}{c|}{708.0 (347.6)} &
  \multicolumn{1}{c|}{317.9 (283.8)} &
  \multicolumn{1}{c|}{786.5 (300.1)} &
  782.1(323.5 \\ \cline{2-14} 
 &
  10 &
  \multicolumn{1}{c|}{-} &
  \multicolumn{1}{c|}{648.0 (0.0)} &
  \multicolumn{1}{c|}{-} &
  \multicolumn{1}{c|}{-} &
  \multicolumn{1}{c|}{2168.5 (672.2)} &
  \multicolumn{1}{c|}{2190.9 (678.2)} &
  \multicolumn{1}{c|}{235.3 (109.4)} &
  \multicolumn{1}{c|}{970.4 (342.5)} &
  \multicolumn{1}{c|}{632.9 (306.8)} &
  \multicolumn{1}{c|}{182.8 (194.2)} &
  \multicolumn{1}{c|}{747.3 (281.0)} &
  676.7(233.3) \\ \hline

\end{tabular}%
}
\vspace{-10pt}
\end{table*}

\begin{table*}[t]
\captionsetup{skip=0pt}
\caption{Makespan and mean timestep of symmetric layouts (standard deviations in the parentheses)}
\label{tab:symmetric_table}
\centering
\resizebox{2.0\columnwidth}{!}{%
\begin{tabular}{|c|c|cccccccccccc|}
\hline
\multirow{2}{*}{Env.} &
  \multirow{2}{*}{\#robot} &
   
  \multicolumn{6}{c|}{Makespan} &
  \multicolumn{6}{c|}{Mean timestep} \\ \cline{3-14} 
 &
   &
  \multicolumn{1}{c|}{ORCA} &
  \multicolumn{1}{c|}{APF} &
  \multicolumn{1}{c|}{$\mu$NAV} &
  \multicolumn{1}{c|}{RPF} &
  \multicolumn{1}{c|}{APF-RS} &
  \multicolumn{1}{c|}{APF-LS} &
  \multicolumn{1}{c|}{OCRA} &
  \multicolumn{1}{c|}{APF} &
  \multicolumn{1}{c|}{$\mu$NAV} &
  \multicolumn{1}{c|}{RPF} &
  \multicolumn{1}{c|}{APF-RS} &
  APF-LS \\ \hline
\textsf{Flat} &
  2 &
  \multicolumn{1}{c|}{1033.6 (177.3} &
  \multicolumn{1}{c|}{-} &
  \multicolumn{1}{c|}{522.8 (123.5)} &
  \multicolumn{1}{c|}{679.9 (122.2)} &
  \multicolumn{1}{c|}{499.5 (83.7)} &
  \multicolumn{1}{c|}{532.3 (145.5)} &
  \multicolumn{1}{c|}{1027.3 (187.9)} &
  \multicolumn{1}{c|}{-} &
  \multicolumn{1}{c|}{490.8 (131.6)} &
  \multicolumn{1}{c|}{648.8 (138.6)} &
  \multicolumn{1}{c|}{494.0 (83.1} &
  512.9(126.3) \\ \hline
\textsf{Cylind} &
  2 &
  \multicolumn{1}{c|}{-} &
  \multicolumn{1}{c|}{-} &
  \multicolumn{1}{c|}{501.0 (62.7)} &
  \multicolumn{1}{c|}{-} &
  \multicolumn{1}{c|}{825.2 (245.4)} &
  \multicolumn{1}{c|}{838.1 (241.0)} &
  \multicolumn{1}{c|}{-} &
  \multicolumn{1}{c|}{-} &
  \multicolumn{1}{c|}{472.2 (82.1)} &
  \multicolumn{1}{c|}{-} &
  \multicolumn{1}{c|}{784.4 (235.5)} &
  785.4(230.8) \\ \hline
\multirow{3}{*}{\textsf{Swap}} &
  6 &
  \multicolumn{1}{c|}{233.6 (10.7)} &
  \multicolumn{1}{c|}{260.2 (19.7)} &
  \multicolumn{1}{c|}{601.0 (82.9)} &
  \multicolumn{1}{c|}{432.7 (30.2)} &
  \multicolumn{1}{c|}{210.8 (41.7)} &
  \multicolumn{1}{c|}{210.4 (22.5)} &
  \multicolumn{1}{c|}{233.5 (10.7)} &
  \multicolumn{1}{c|}{241.9 (15.3)} &
  \multicolumn{1}{c|}{324.1 (83.7)} &
  \multicolumn{1}{c|}{323.6 (13.3)} &
  \multicolumn{1}{c|}{198.3 (38.0} &
  196.0(13.7) \\ \cline{2-14} 
 &
  8 &
  \multicolumn{1}{c|}{357.9 (64.6)} &
  \multicolumn{1}{c|}{264.9 (20.0)} &
  \multicolumn{1}{c|}{-} &
  \multicolumn{1}{c|}{570.4 (50.9)} &
  \multicolumn{1}{c|}{276.1 (60.7)} &
  \multicolumn{1}{c|}{268.5 (30.6)} &
  \multicolumn{1}{c|}{357.6 (64.1)} &
  \multicolumn{1}{c|}{241.3 (17.2)} &
  \multicolumn{1}{c|}{287.0 (94.5)} &
  \multicolumn{1}{c|}{382.6 (17.5)} &
  \multicolumn{1}{c|}{245.8 (64.7)} &
  238.9(20.7) \\ \cline{2-14} 
 &
  10 &
  \multicolumn{1}{c|}{685.0 (73.9)} &
  \multicolumn{1}{c|}{261.5 (19.8)} &
  \multicolumn{1}{c|}{-} &
  \multicolumn{1}{c|}{-} &
  \multicolumn{1}{c|}{310.6 (72.6)} &
  \multicolumn{1}{c|}{256.7 (29.4)} &
  \multicolumn{1}{c|}{680.4 (73.2)} &
  \multicolumn{1}{c|}{230.8 (15.0)} &
  \multicolumn{1}{c|}{296.6 (87.5)} &
  \multicolumn{1}{c|}{569.4 (32.5)} &
  \multicolumn{1}{c|}{260.1 (67.1)} &
  223.0(16.6) \\ \hline
\end{tabular}%
}\vspace{-10pt}

\end{table*}

\subsubsection{Symmetric layouts}
\begin{figure}
\vspace{-5pt}
\captionsetup{skip = 0pt}
    \centering
    \begin{subfigure}[b]{\columnwidth}
        \centering
        \captionsetup{skip = -7pt}
        \includegraphics[width=\textwidth]{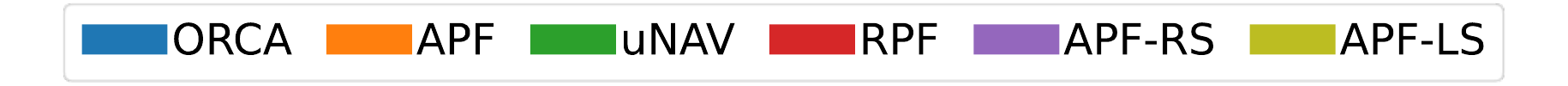}
        \label{fig:legend}
    \hfill \vspace{-13pt}
    \end{subfigure}
    \begin{subfigure}[b]{0.49\columnwidth}
        \centering
        \captionsetup{skip = -5pt}
        \includegraphics[width=\textwidth]{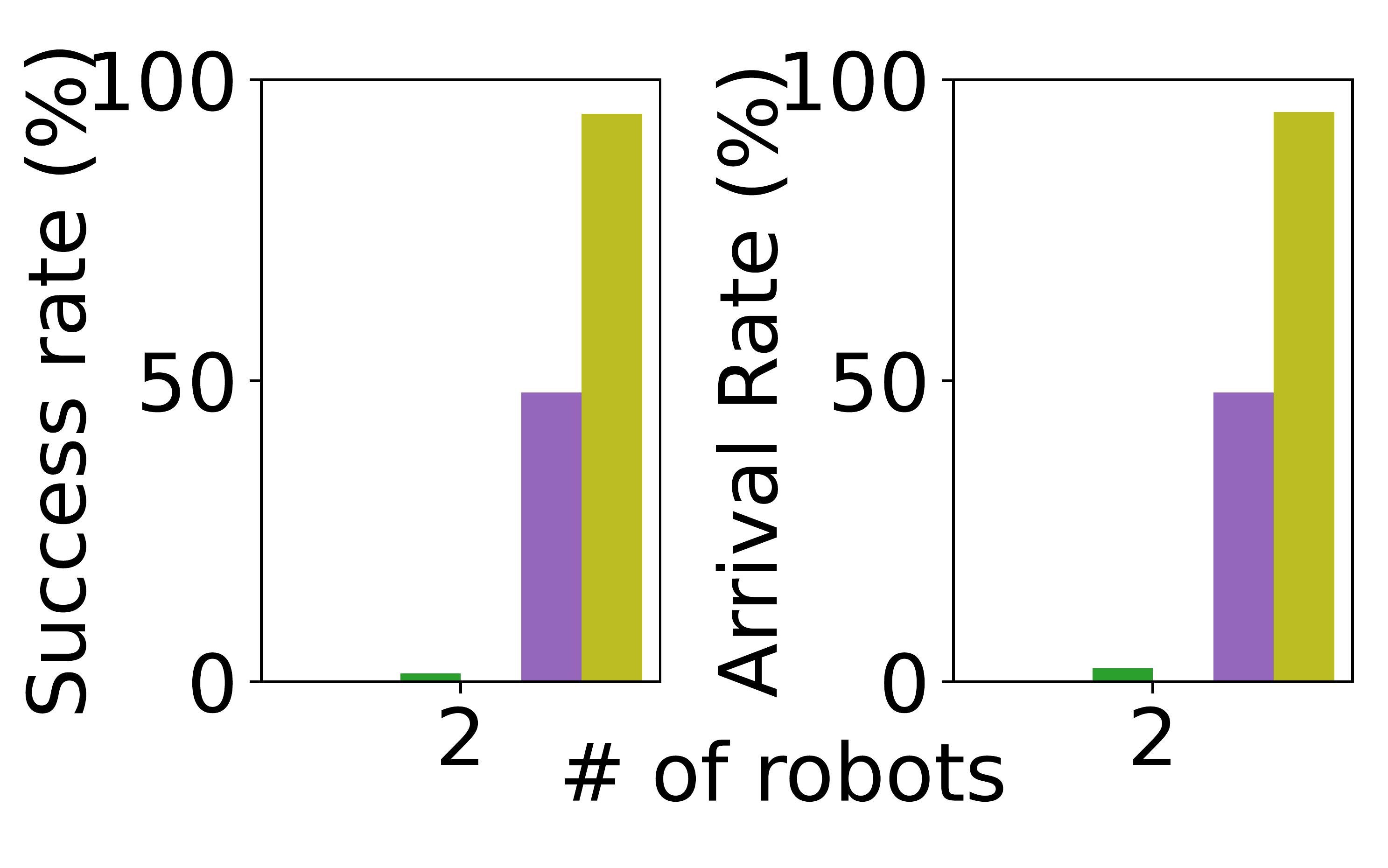}
        \caption{Result of \textsf{Flat}}
        \label{fig:flat_test_results}
    \end{subfigure}
    \begin{subfigure}[b]{0.49\columnwidth}
        \centering
        \captionsetup{skip = -5pt}
        \includegraphics[width=\textwidth]{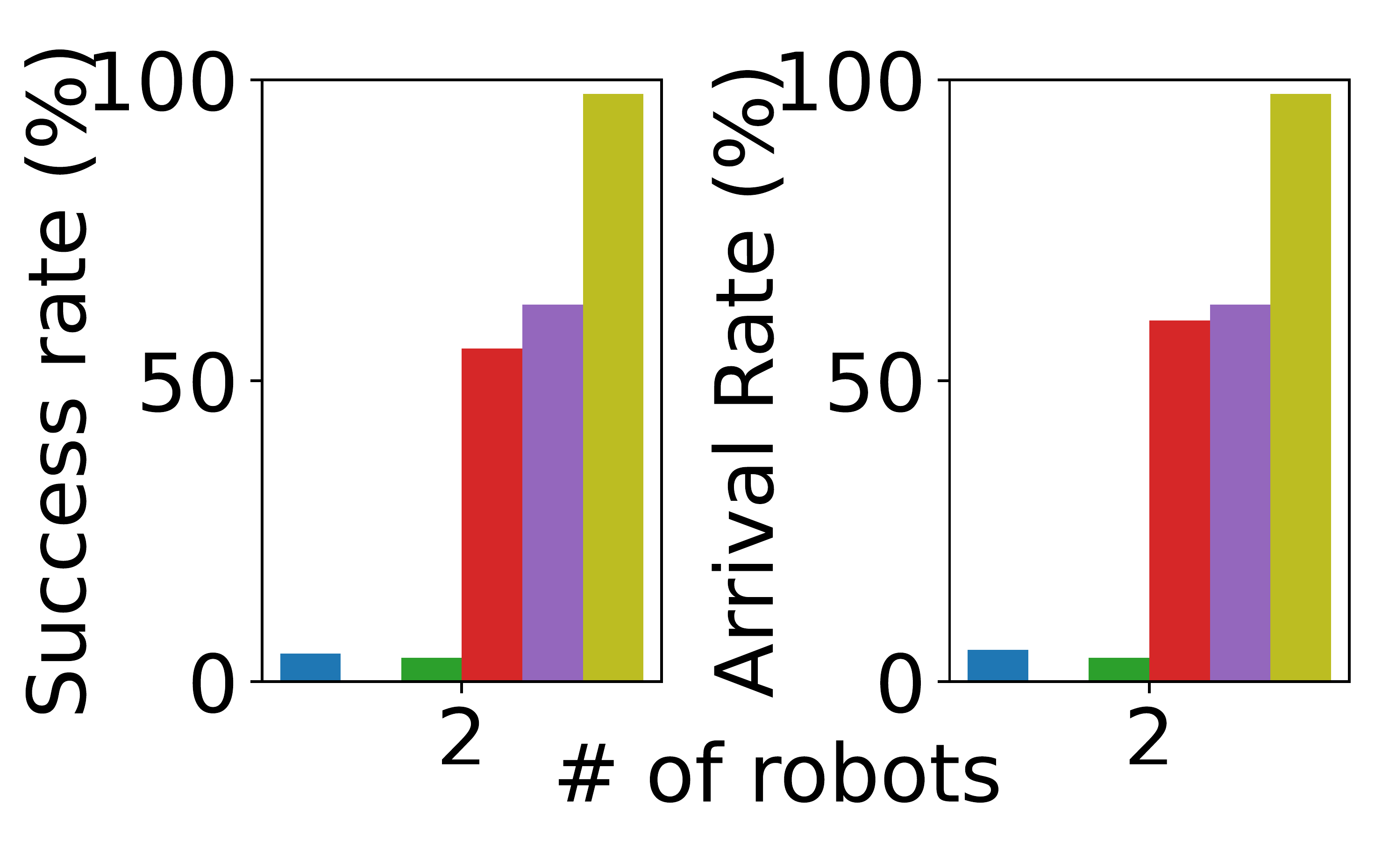}
        \caption{Result of \textsf{Cylind}}
        \label{fig:cylinder_test_results}
    \end{subfigure}
    \begin{subfigure}[b]{0.98\columnwidth}
        \centering
        \captionsetup{skip = -6pt}
        \includegraphics[width=\textwidth]{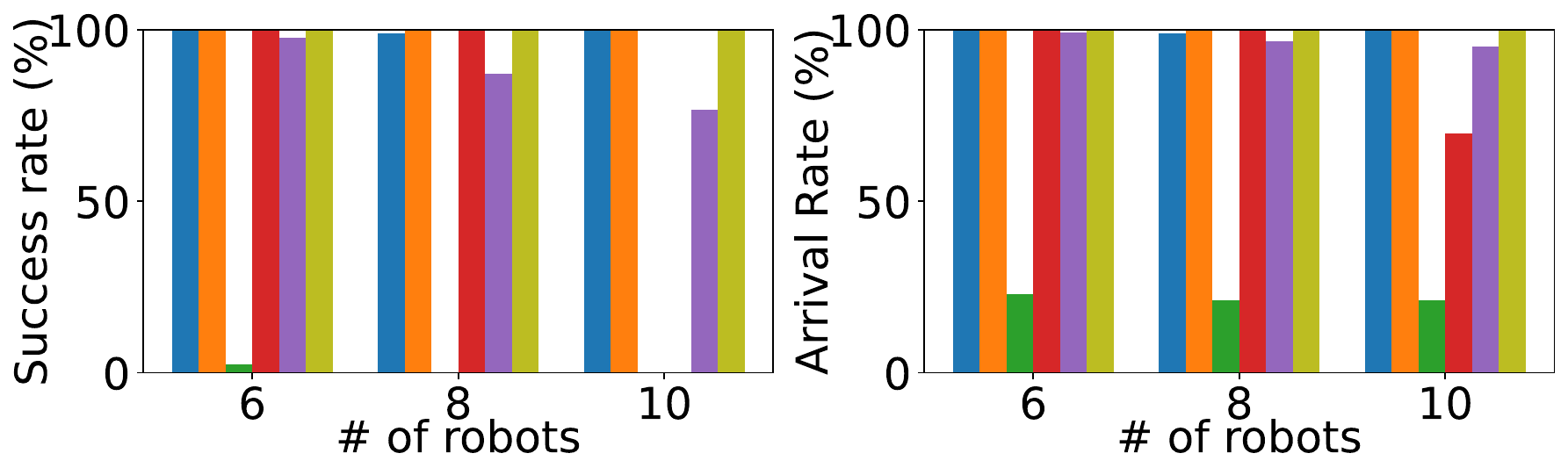}
        \caption{Result of \textsf{Swap}}
        \label{fig:posswap_test_results}
    \end{subfigure}
    \caption{Success and arrival rates of symmetric layouts} \vspace{-5pt}
    \label{fig:symmetric_test_results}
\end{figure}

Fig.~\ref{fig:symmetric_test_results} shows success and arrival rates while makespan and mean timestep are shown in Table~\ref{tab:symmetric_table}. The time limit $T$ for \textsf{Flat}, \textsf{Cylind}, and \textsf{Swap} are $1500$, $1500$, and $1000$, respectively. In \textsf{Flat} and \textsf{Cylind}, LS outperforms RS by $35\%$ and $48\%$, respectively, while the baselines show the performance much below than LS. ORCA and APF fail in \textsf{Flat} and \textsf{Cylind} owing to deadlocks and local minima at the nonconvex dents, respectively.  $\mu$NAV consistently experiences the false-WF as it focuses only on the tangent direction of the nearest obstacle, which makes the robots often misinterpret other robots as walls. RPF handles such situations better in \textsf{Flat} but fails entirely in \textsf{Cylind}. Similar to the case with the real-world layouts, the makespan and mean timestep of RS and LS are higher than some baseline methods owing to the low success rates of the compared ones. In \textsf{Swap}, the success rate of RS declines while LS achieves 100\%. RPF succeeds in tests with up to $8$ robots, but with $10$, excessive collision avoiding moves leads to failure. As the number of robots increased, ORCA and APF experience near-deadlock situations, leading to increased makespan and mean timestep values. The result demonstrates that our LS method is effective in mitigating multi-robot conflicts, such as the false-WF.

\subsection{Physical Robot Experiments}
\vspace{-3pt}

\begin{figure}
    \captionsetup{skip = 1pt}
    \centering
    \includegraphics[width=\columnwidth]{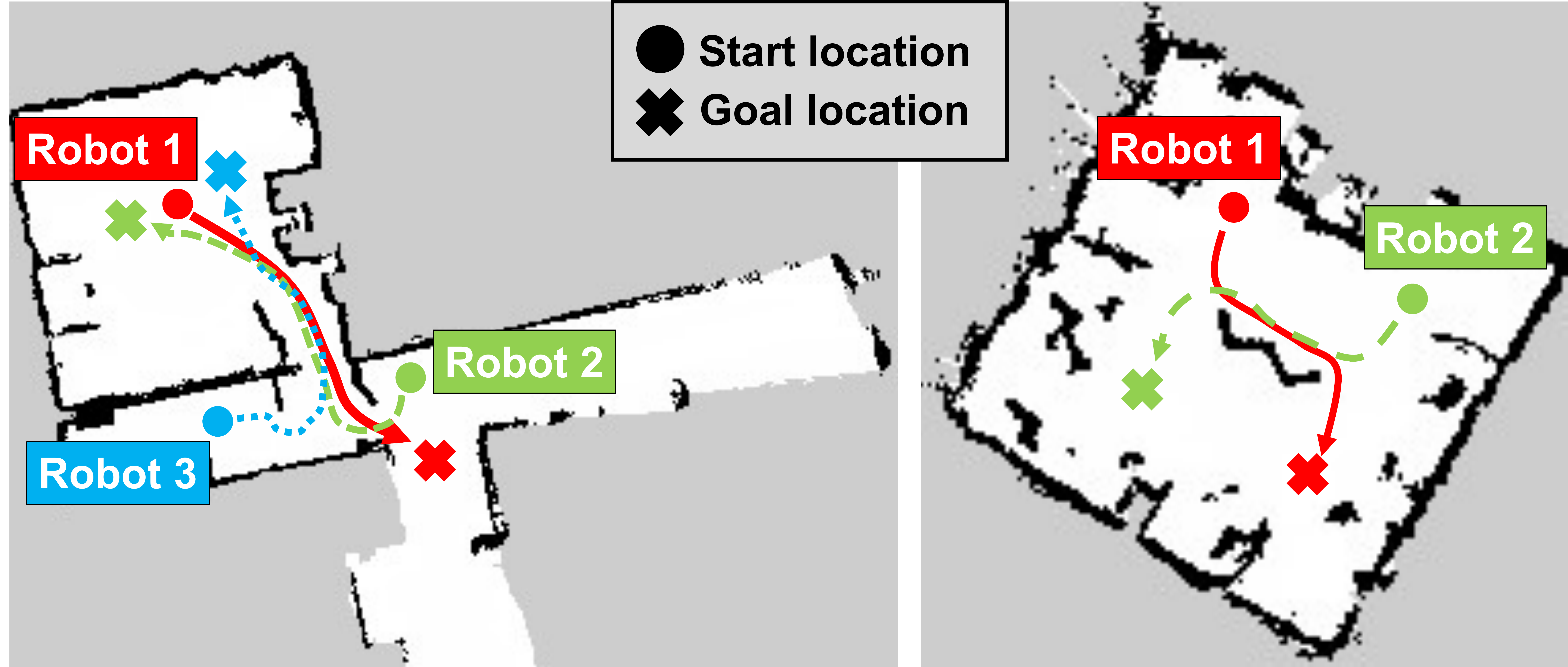}
    \caption{Illustration of two real-robot tasks. (Left) Task 1: three robots navigate an office environment with nonconvex obstacles. (Right) Task 2: two robots on opposite sides of a wall.}
    \label{fig:real_robot_test} \vspace{-15pt}
\end{figure}

We demonstrated the effectiveness of our proposed method in two real-robot tasks. In Task 1 (Fig.~\ref{fig:real_robot_test} left), three robots navigated through an office environment with nonconvex obstacles, where our RS and LS successfully escape local minima, unlike the vanilla APF. In Task 2 (Fig.~\ref{fig:real_robot_test} right), two robots positioned symmetrically on opposite sides of a wall face deadlocks with vanilla APF, but both RS and LS completed the task. These results show that our method enhances navigation in real-world multi-robot systems.

\section{Conclusion}
\vspace{-3pt}

We propose a decentralized, mapless multi-robot navigation method combining APF and WF to overcome local minima in nonconvex environments. Using rule-based and learning-based switching, our approach enables collision-free navigation with local sensors and no communication. Experiments show our methods outperform baselines, with LS excelling in confined spaces and reducing multi-robot conflicts. Future work will expand physical robot tests in more real-world scenarios.


\clearpage
\bibliographystyle{IEEEtran}
\bibliography{references}

\end{document}